%% file: main.tex
\documentclass[letterpaper, 10 pt, conference]{ieeeconf}  

\IEEEoverridecommandlockouts                              

\usepackage{cite}
\usepackage{amsmath,amssymb,amsfonts}

\usepackage[hidelinks]{hyperref}

\usepackage{cleveref}

\usepackage{algorithmic}
\usepackage{graphicx}
\usepackage{textcomp}
\usepackage{xcolor}
\usepackage{multirow}
\usepackage{physics}
\usepackage[para]{threeparttable}
\usepackage[inkscapeformat=png]{svg}
\usepackage{tabularx}
\usepackage{float}
\usepackage{soul}
\usepackage{gensymb}

\usepackage{glossaries}
\usepackage{subfiles}
\subfile{acronyms}

\title{\LARGE \bf
Compact robotic gripper with tandem actuation for selective fruit harvesting
}

\author{Alejandro Velasquez$^{1}$, Cindy Grimm$^{1}$, Joseph R. Davidson$^{1}$ 
\thanks{This research is supported in part by USDA-NIFA through the Cyber Physical Systems program (Award No. 2020-67021-31525) and NSF RI: Small: Leveraging  Human  Manipulation Skills to Advance Near Contact Robotic Grasping and In-Hand Stabilization (Award No. 1911050).}
\thanks{$^1$Collaborative Robotics and Intelligent Systems (CoRIS) Institute, Oregon State University, Corvallis OR 97331, USA {\tt\footnotesize \{velasale, cindy.grimm, joseph.davidson\}@oregonstate.edu}}
}

\begin{document}

\maketitle
\thispagestyle{empty}
\pagestyle{empty}

\begin{abstract}
Selective fruit harvesting is a challenging manipulation problem due to occlusions and clutter arising from plant foliage. 
A harvesting gripper should i) have a small cross-section, to avoid collisions while approaching the fruit; ii) have a soft and compliant grasp to adapt to different fruit geometry and avoid bruising it; and iii) be capable of rigidly holding the fruit tightly enough to counteract detachment forces.
Previous work on fruit harvesting has primarily focused on using grippers with a single actuation mode, either suction or fingers. 
In this paper we present a compact robotic gripper that combines the benefits of both. 
The gripper first uses an array of compliant suction cups to gently attach to the fruit. After attachment, telescoping cam-driven fingers deploy, sweeping obstacles away before pivoting inwards to provide a secure grip on the fruit for picking. 
We present and analyze the finger design for both ability to sweep clutter and maintain a tight grasp. Specifically, we use a motorized test bed to measure grasp strength for each actuation mode (suction, fingers, or both). We apply a tensile force at different angles ($\mathbf{0}$\textdegree, $\mathbf{15}$\textdegree, $\mathbf{30}$\textdegree~and $\mathbf{45}$\textdegree), and vary the point of contact between the fingers and the fruit. We observed that with both modes the grasp strength is approximately $\mathbf{40~N}$. We use an apple proxy to test the gripper's ability to obtain a grasp in the presence of occluding apples and leaves, achieving a grasp success rate over $\mathbf{96\%}$ (with an ideal controller). Finally, we validate our gripper in a commercial apple orchard.

\end{abstract}

\section{Introduction}
\input{1_Introduction}

\section{Related Work}
\input{2_Related_Work}

\section{Gripper Design}
\input{3_Gripper_Design}

\section{Evaluation and Validation}
\input{4_Methodology}

\section{Results and Discussion}
\input{5_Results}

\section{Conclusion}
\input{6_Conclusion}

\section*{ACKNOWLEDGMENT}
Funded in part by NSF grant 2024872 and USDA grant 2020-01461. The primary author thanks the Fulbright-Colombia and Minciencias-Colombia for their financial support. We also thank Allan Bros., Inc. for supporting our field experiments in their commercial orchard.

\bibliographystyle{unsrt}
\bibliography{main}
\end{document}

%% file: acronyms.tex
\makeglossaries

\newacronym{alt}{ALT}{Aggregated Linear Trend}
\newacronym{auc}{AUC}{Area Under the Curve}
\newacronym{cs}{CS}{Configuration Space}
\newacronym{dl}{DL}{Deep Learning}
\newacronym{ds}{DS}{Dynamical System}
\newacronym{dof}{DOF}{Degrees Of Freedom}
\newacronym{eef}{EEF}{End Effector}
\newacronym{fea}{FEA}{Fluidic Elastomeric Actuators}
\newacronym{fov}{FOV}{Field Of View}
\newacronym{fdf}{FDF}{Fruit Detachment Force}
\newacronym{gmr}{GMR}{Gaussian Mixture Regression}
\newacronym{ihc}{IHC}{in-hand camera}
\newacronym{ik}{IK}{Inverse Kinematics}
\newacronym{gmm}{GMM}{Gaussian Mixture Model}
\newacronym{lfd}{LFD}{Learning From Demonstrations}
\newacronym{lstm}{LSTM}{Long Short Term Memory}
\newacronym{ml}{ML}{Machine Learning}
\newacronym{ode}{ODE}{Ordinary Differential Equation}
\newacronym{rl}{RL}{Reinforcement Learning}
\newacronym{rfc}{RFC}{Random Forest Classifier}
\newacronym{rnn}{RNN}{Recurrent Neural Network}
\newacronym{sfh}{SFH}{Selective Fruit Harvesting}
\newacronym{sma}{SMA}{Shape Memory Alloy}
\newacronym{tcp}{TCP}{Tool Center Point}
\newacronym{tof}{TOF}{time of flight}

%% file: 1_Introduction.tex




Most fresh market fruits and vegetables are picked by the human hand. In many parts of the world, rising costs and labor shortages threaten the long-term economic sustainability of manually harvesting these specialty crops~\cite{van_henten_greenhouse_2006, tong_understanding_nodate, Zhang2020a, kootstra_selective_2021, fang_fruit_2023}. While there has been decades of research on robotic harvesting and tremendous advances in computing, sensors, and deep learning, there are few, if any, robotic harvesting systems that are commercially available to growers. Selective harvesting has proven to be a difficult manipulation problem.    

There are several challenges with robotic harvesting. First, the environment is very unstructured. In the plant canopy there are heavy occlusions, clutter, and clusters (i.e. touching fruit). Secondly, depending on the cultivation system and crop, the mechanics of the environment can be highly variable. There can be soft vegetation like leaves interspersed with rigid structures such as limbs, trellis wires, and irrigation lines. Additionally, fruit are free-floating and swing away when pushed. A third challenge is that while fruit can often be easily damaged, they still have to be manipulated in such a way as to deliver significant detachment forces. For example, prior work in apples shows that the fruit detachment force varies with the picking pattern~\cite{Li2016, Davidson2016, ji_new_2024, tong_understanding_nodate}. Linear pulling results in the highest detachment forces. Other patterns (e.g. rotate, bend and pull) may lead to lower forces and avoid damage to the fruit and tree, but to perform this type of pick the gripper must be able to hold on to the fruit in the presence of detachment forces coming from various directions.





\begin{figure}[!t]
      \centering      
      \includegraphics[width=1.0\linewidth]{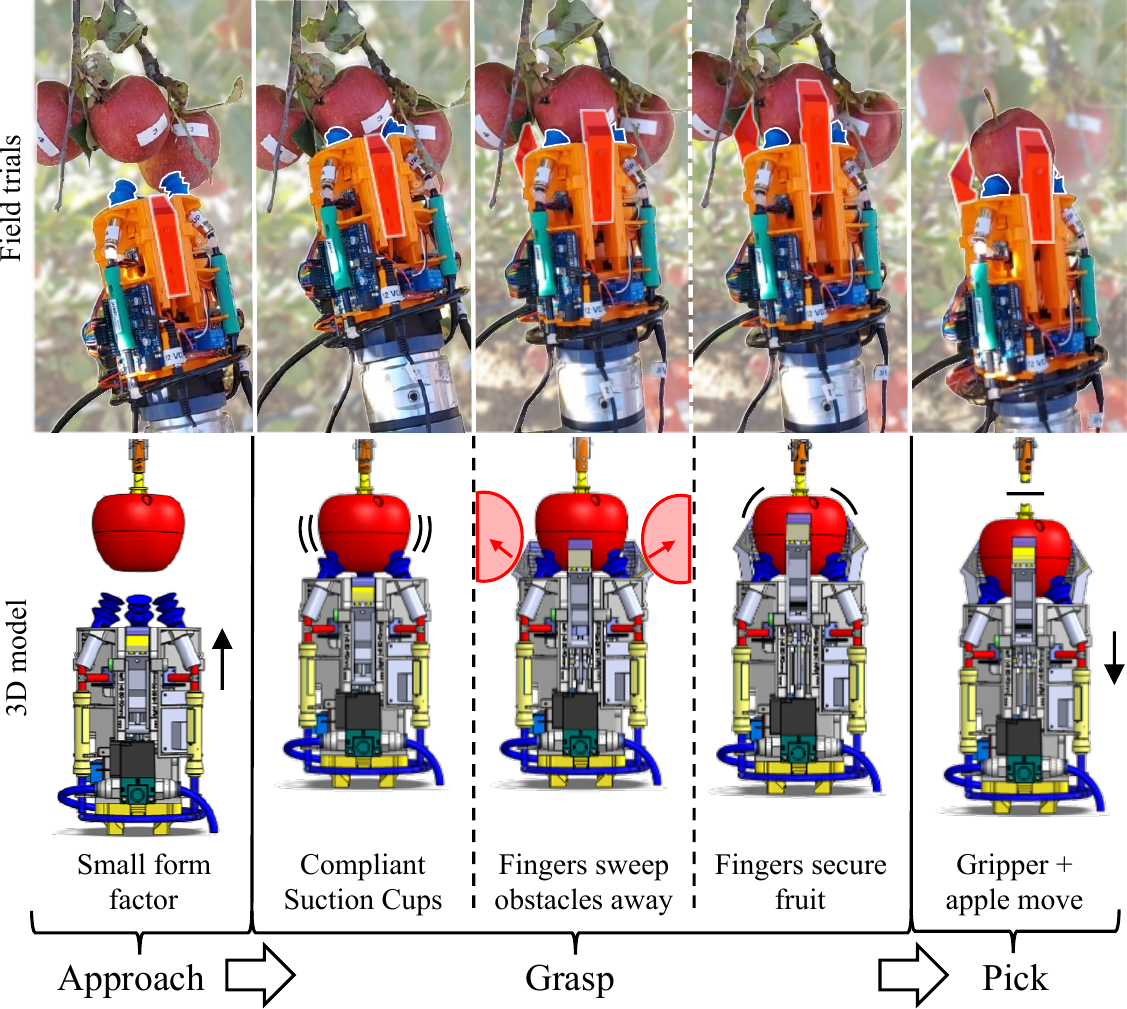}
      \caption{From left to right. \textit{Approach}: The gripper approaches the fruit with a small form factor to reduce collisions. \textit{Grasp}: An initial compliant grasp with suction cups is followed by a grasp with cam-driven fingers that wedge between neighboring fruit. \textit{Pick}: The apple is secured in the gripper during the picking motion.}
      \label{fig:abstractImage}
\end{figure}

Meeting all of these challenges with a single robotic gripper is difficult. Most harvesting grippers integrate a single actuation mode such as suction, fingers, or jamming~\cite{elfferich_soft_2022}. However, each of these actuation modes has limitations. While suction-based designs tend to have a small form factor, they can require significant vacuum to generate the required detachment forces. They are also less robust to the shear forces that occur in picking motions that are not purely linear. With fingers, the problems are the robot's approach to the target fruit and its initial grasp. A gripper with a wide opening span is likely to have unintended collisions in the cluttered canopy that could lead to picking failures. \textit{Our approach combines both actuation modes -- suction and fingers -- in a single design}.  

 We present a robotic gripper (Fig.~\ref{fig:abstractImage}) that uses \textit{tandem actuation} to address the conflicting challenges of selective fruit harvesting. The gripper includes an array of three small, compliant suction cups which attach to the fruit. This array has a small form factor (a cross section similar to the size of the fruit) to minimize collisions, and the compliance plus suction minimize the likelihood that the fruit will be pushed away~\cite{Velasquez2024}. After suction is achieved, three telescoping, cam-driven fingers then extend outwards to grip the fruit. We designed the finger paths to first create a sweeping motion that pushes away neighboring obstacles that could interfere with the grasp, such as fruit and vegetation. The fingers then sweep in to ensure a stable grasp during the picking motion. Unlike articulated fingers, this grasp is primarily maintained through mechanical advantage, rather than motor activation. While we designed our gripper for apple harvesting, the concept of tandem actuation with suction and telescoping/sweeping fingers could be extended to other types of fruits and vegetables. 
 
After reviewing the prior work, we begin by presenting the gripper's design along with a static model for analyzing the device's transmission ratio in `finger-mode' (the suction cup design was presented in~\cite{Velasquez2024}). We then experimentally evaluate the gripper's grasp strength for varying apple pose and disturbance forces, observing that strength is substantially higher with our tandem actuation (approximately $40~N$) and well above the expected fruit detachment force. During lab picking trials with a robot manipulator and our physical apple proxy~\cite{Velasquez2022}, we show that the telescoping, sweeping finger design is successful at establishing a secure grasp in clutter (i.e. clusters and leaves). Finally, we validate our gripper in a commercial apple orchard (using an ideal controller and a pull-back picking motion). In summary, our contributions are the following:
\begin{enumerate}
    \item A dual actuation gripper that embodies two behaviors: i) compliant attachment with the fruit via suction; ii) secure, stable grasping of the fruit using fingers.
    \item A cam-driven finger mechanism with two deployment stages. The first stage uses the wedge-shaped finger pads to sweep obstacles away from the target fruit and the second stage clamps the fruit, using mechanical advantage to maintain that grasp. 
    \item An assessment protocol to evaluate a gripper's robustness to variations in kinematic pose and dynamic forces.
\end{enumerate}

%% file: 2_Related_Work.tex


The recent prior work on robotic grippers for selective fruit harvesting agrees on the need for mechanical compliance. Here we review the state-of-the-art in compliant grippers, variable stiffness grippers, and tandem actuation, focusing on the literature most relevant to fruit harvesting.


\subsection{Compliant grippers}
Compliant grippers are well-suited for manipulating fruit in cluttered plant canopies. Compliance can be created by using a passive approach with mechanical design (e.g. springs) and soft materials (e.g. silicone parts), or through an active approach with sensors and feedback control. Elfferich et al. \cite{elfferich_soft_2022} provide a complete survey of soft robotic grippers used for crop handling or harvesting. One significant finding is that $56$ out of the $78$ studied grippers were meant for pick and place operations, but not for fruit detachment in the field. 
In Lie et al.~\cite{liu_topology_2018}, the authors used topology optimization to develop a compliant mechanism to grasp objects with sizes between $42~mm$ and $141~mm$ and payloads up to $2~kg$.
In another work, Gunderman et. al. \cite{gunderman_tendon-driven_2022} developed a tendon-driven soft gripper for blackberry harvesting, small and soft fruits that require low picking forces ($<2N$).  
In some cases \cite{xu_compliant_2021, chi_universal_2024}, compliance is wisely used twofold: i) to adapt to the targeted objects, and ii) to measure the intrinsic force from gripper deformation. 

\subsection{Variable stiffness grippers}
Variable stiffness grippers can be created using techniques such as particle jamming, `tunable' springs, smart materials (e.g. shape memory alloys), etc. A main advantage of variable stiffness is the ability to modify the gripper's behavior from compliant and adaptive to stiff and rigid. The universal gripper~\cite{brown_universal_2010} leverages the stiffness variation obtained when granular material is under different air pressures. Its initial compliance allows for good adaptations to objects with different sizes and shapes. However, particle jamming is not suitable for fruit harvesting for two reasons. First, the targeted object needs to rest on a support surface (i.e. not be suspended in the air). Second, curved, convex geometries may require an infeasibly large interface with the jamming surface to obtain a good grip. In a different approach to variable stiffness, Goshtasbi et al. \cite{goshtasbi_bioinspired_2023} develop a suction cup-based gripper inspired by Octopus suckers. The key novelty is the neck of the suction cups which can be adjusted with different stiffnesses. The necks are compliant before grasping the object, and then turn rigid. This suction cup-concept has a lot of potential for applications like ours. However, one caveat is that some fruits require high detachment forces that small suction cups can't deliver. We address this problem by combining small suctions cups with fingers. 

\begin{figure*}[!htp]
  \centering      
  \includegraphics[width=0.85\linewidth]{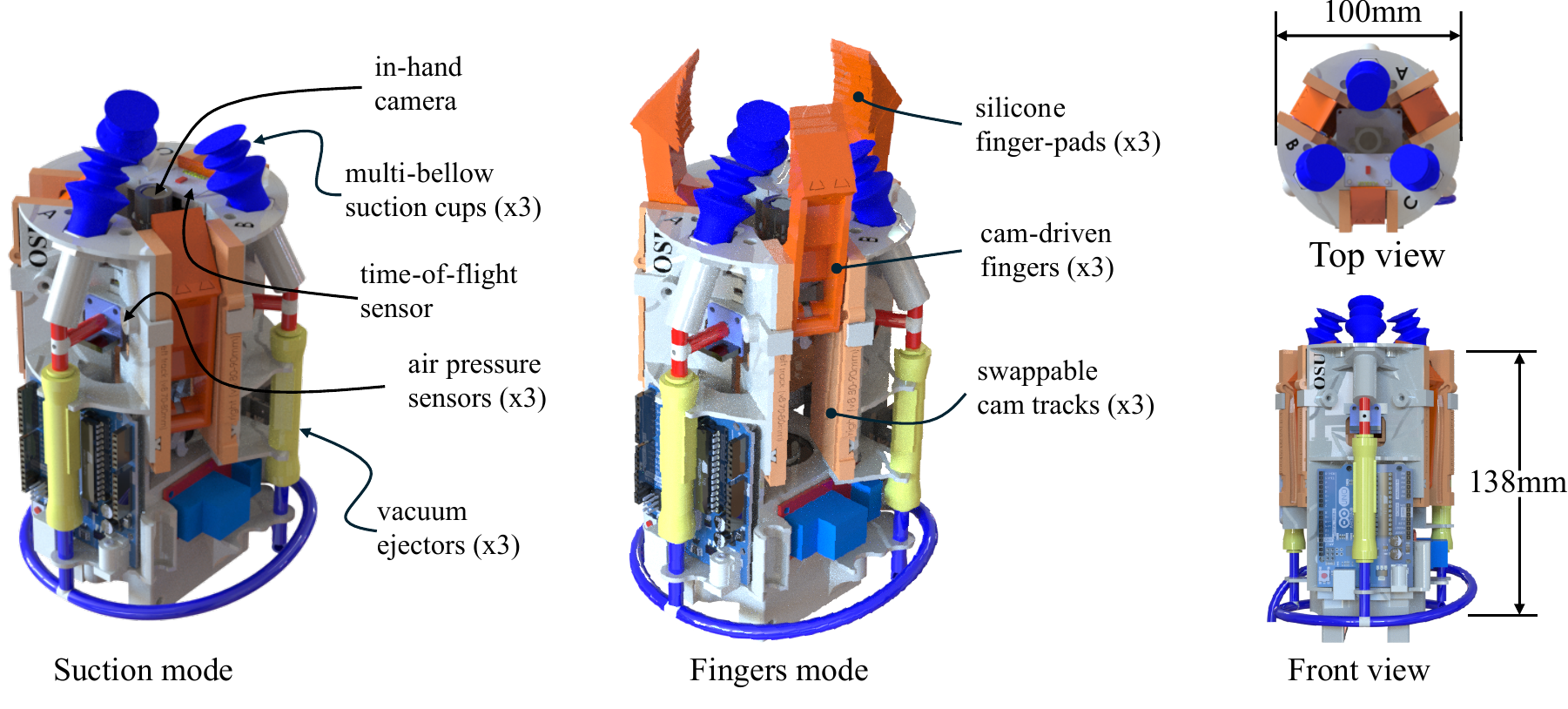}
  \caption{Gripper render. \textit{Left}: Suction mode~\cite{Velasquez2024} with fingers retracted and in-hand perception components labeled. \textit{Middle}: Gripper with fingers deployed. \textit{Top and bottom right}: Gripper's main dimensions with fingers retracted.}
  \label{fig:des_render}
\end{figure*}

\subsection{Tandem actuation grippers}
A promising trend in gripper design is combining different actuation modes. The most common approach is the integration of fingers with a single suction cup on the gripper's palm. Some of these cases use rigid linkage driven fingers with rubber pads~\cite{kang_design_2019, 10306271, ronzhin_model_2020}, whereas other cases use fluidic elastomeric actuator fingers. In another configuration, Bryan et al.~\cite{8914479} used cable driven fingers that had suction cups on the finger pads. With finger-based grippers, the robot will usually approach the target object with the fingers in a wide opening span. This is problematic for tree fruit harvesting due to the highly cluttered workspace. Our design approaches the fruit with the fingers retracted and then extends them after the suctions cups have attached to the fruit. 




%% file: 3_Gripper_Design.tex


We consider design specifications related to the kinematic and dynamic challenges of fruit picking mentioned earlier. We tackle these specifications by combining two actuation modes that complement each other: compliant multi-bellow suction cups and cam-driven fingers. The first provides a robust initial attachment and the latter deploys from underneath the gripper's palm, sweeping obstacles away and securing the apple for further manipulation (i.e. the picking motion).

\begin{figure*}[htp]
  \centering      
  \includegraphics[width=1.0\linewidth]{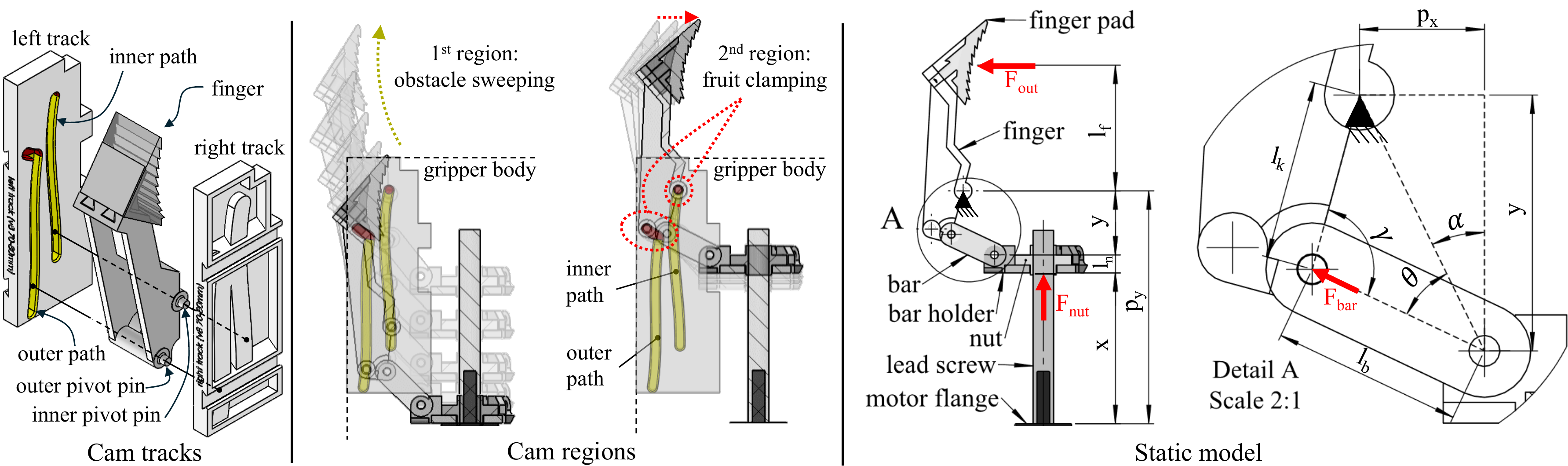}
  \caption{Cam-driven finger mechanism. \textit{Left}: Exploded view of a single finger and cam tracks. The finger's outer pivot pin slides along the outer path of the tracks and the inner pivot pin slides along the inner path. \textit{Middle}: Side view of the resulting path followed by the telescoping fingers -- the fingers first curve outwards (yellow areas of the paths) to sweep obstacles away from the targeted fruit before moving inwards (dotted red encircled areas) to secure the fruit. \textit{Right}: Detail view of linkage at the second region with the corresponding geometric configuration of the force transmission model.}
  \label{fig:des_gripper}
\end{figure*}

\subsection{Design specifications}
Based on our review of the prior art, we considered the following design specifications for the gripper:\begin{itemize}
    \item \textbf{Form factor}: The gripper should have a cross-section similar in shape and size to the targeted fruit to minimize unintended collisions in the cluttered canopy. 
    \item \textbf{In-hand sensing}: The gripper should have in-hand perception such as cameras and time-of-flight sensors to allow active fruit localization during the approach.
    \item \textbf{Grasp strength}: The gripper should: i) provide sufficient grasp strength to withstand the forces required to detach the fruit; ii) maintain a stable grasp of the fruit during twisting motions; and iii) accomplish i) and ii) without bruising the fruit with the grasp itself.
    \item \textbf{Robustness}: The gripper should be able to pick fruit in clusters (i.e. fruit that are touching), fruit adjacent to limbs, and fruit adjacent to other rigid obstacles such as trellis wires and irrigation lines.
\end{itemize}
\subsection{Design description}
\textit{Actuation.} Our gripper employs two actuation modes (\cref{fig:des_render}). The first mode is suction with three compliant bellows suction cups instead of one (described in our previous work \cite{Velasquez2024}). The multiple bellows arrangement has the benefits of increasing the likelihood of fruit attachment, creating an open field of view for perception sensors located in the center of the palm, and providing three contact points to help stabilize the fruit. The second actuation mode, which begins after suction is complete, is telescoping, cam-driven fingers that sweep outward from the gripper and provide a strong grasp to complete the picking motion (described in~\cref{subsub:camfingers}). 


\textit{Sensors.} As described in our previous work~\cite{Velasquez2024}, the gripper palm has an in-hand camera (ELP High Speed USB Camera 1080P) and a time-of-flight sensor (Adafruit VL53L0X) for visual servoing to the apple (the method used for visual servoing is not within scope of this paper). Additionally, we placed air pressure sensors (Adafruit MPRLS) near the intake of each suction cup to determine if the suction cups are engaged.

\subsection{Cam-driven fingers}
\label{subsub:camfingers}
The primary motivation for the cam-driven mechanism is to create two different finger `behaviors' with the same mechanism: \textit{obstacle sweeping} and \textit{fruit clamping}. The geometry of the cam paths defines the finger behaviors. Each finger has two pivot pins (inner and outer) that slide along the cam paths and define the finger pose (\cref{fig:des_gripper}-\textit{Left}). Cam paths are embedded into modular 3D-printed parts that can be easily replaced for harvesting different types of fruit and fruit sizes  (\cref{fig:des_gripper}-\textit{Left}). The actuator for the three coupled fingers is a stepper motor with a lead screw transmission.  

\textit{First region: obstacle sweeping.} This region of the path prioritizes moving the fingers along the sides of the fruit, following a curved path from underneath the gripper palm and up to the fruit equator (\cref{fig:des_gripper}-\textit{Middle-left}). During this motion, the fingers sweep obstacles (e.g., neighboring fruit, vegetation) away from the fruit with the outer side of wedge-shaped finger-pads. 

\textit{Second region: fruit clamping.} The fingers secure the fruit when the inner pivot pin reaches the end of the inner path (i.e. is constrained by a hard stop), and the outer pivot pin transitions to the last section of the outer path. This results in a rotation of the finger w.r.t. the inner pivot pin, securing the fruit with the inner side of the finger pads (\cref{fig:des_gripper}-\textit{Middle}). The finger pads are made from soft silicone (Dragon Skin\texttrademark~20, Smooth-on Inc., USA) to avoid bruising the fruit and to increase friction. This region of the cam path prioritizes securing the fruit in a tight grasp that can resist pull and twist forces. 

The static model of the fruit clamping region is derived from \cref{fig:des_gripper}-\textit{Right}. Equations \ref{eq:distance} to \ref{eq:ratio}, which use the laws of cosines and sines, describe the force transmission from the \textit{nut} ($F_{nut}$), through the \textit{bar} ($F_{bar}$), up to the \textit{finger-pad} ($F_{out}$). The input to the model is the linear distance $x$ traveled by the lead-screw \textit{nut}. The crank-slider linkage parameters that we used are $p_x=12~mm$, $l_b=18.5~mm$, $l_k=17.5~mm$, $l_f=48~mm$, $p_y=90~mm$, and $l_n=7~mm$.

\begin{equation}
    y = \left({p_y-l_n-x}\right)
\label{eq:distance}
\end{equation}

\begin{equation}
\gamma = \cos^{-1}\left(\frac{{{l_b}^2 + {l_k}^2 - {p_x}^2 - y^2}}{{2l_bl_k}}\right)
\label{eq:gamma}
\end{equation}

\begin{equation}
    \alpha = \tan^{-1}\left(\frac{p_x}{y}\right)
\label{eq:alpha}
\end{equation}

\begin{equation}
    \theta = \sin^{-1}\left(\frac{{l_k \cdot \sin \gamma}}{{\sqrt{p_x^2 + y^2}}}\right)
\label{eq:theta}
\end{equation}

\begin{equation}
F_{\text{bar}} = \frac{F_{\text{nut}}}{\cos(\alpha + \theta)}
\label{eq:Fbar}
\end{equation}

\begin{equation}
{ratio} = \frac{F_{\text{out}}}{F_{\text{nut}}} = \frac{l_k}{l_f} \cdot \frac{\sin\gamma}{\cos(\alpha+\theta)}
\label{eq:ratio}
\end{equation}


Moreover, equations~\cref{eq:wedgeL} to~\cref{eq:Torque} describe the power transmission from the \textit{motor} ($T_{motor}$) to the \textit{nut} ($F_{nut}$)~\cite{shigley2021mechanical}. Power transmission is obtained through a $Tr8x8$ lead screw with the following parameters: $pitch=2~mm$, number of starts $n_s=4$, acme thread angle $\varphi=14.5$\textdegree, external diameter $d_s=8~mm$, and friction coefficient $\mu=0.2$.

\begin{equation}
    l = {pitch} \cdot n_s
\label{eq:wedgeL}
\end{equation}


\begin{equation}
d_{m} = d_s - \frac{pitch}{2}
\label{eq:meanDiameter}
\end{equation}


\begin{equation}
T_{\text{motor}} = \frac{{F_{\text{nut}} \cdot d_{\text{m}}}}{2} \cdot \left(\frac{{l + \pi d_{\text{m}} \mu \sec\varphi}}{{\pi d_{\text{m}} - \mu l \sec\varphi}}\right)
\label{eq:Torque}
\end{equation}

\begin{figure}[!tp]
  \centering      
  \includegraphics[width=0.95\linewidth]{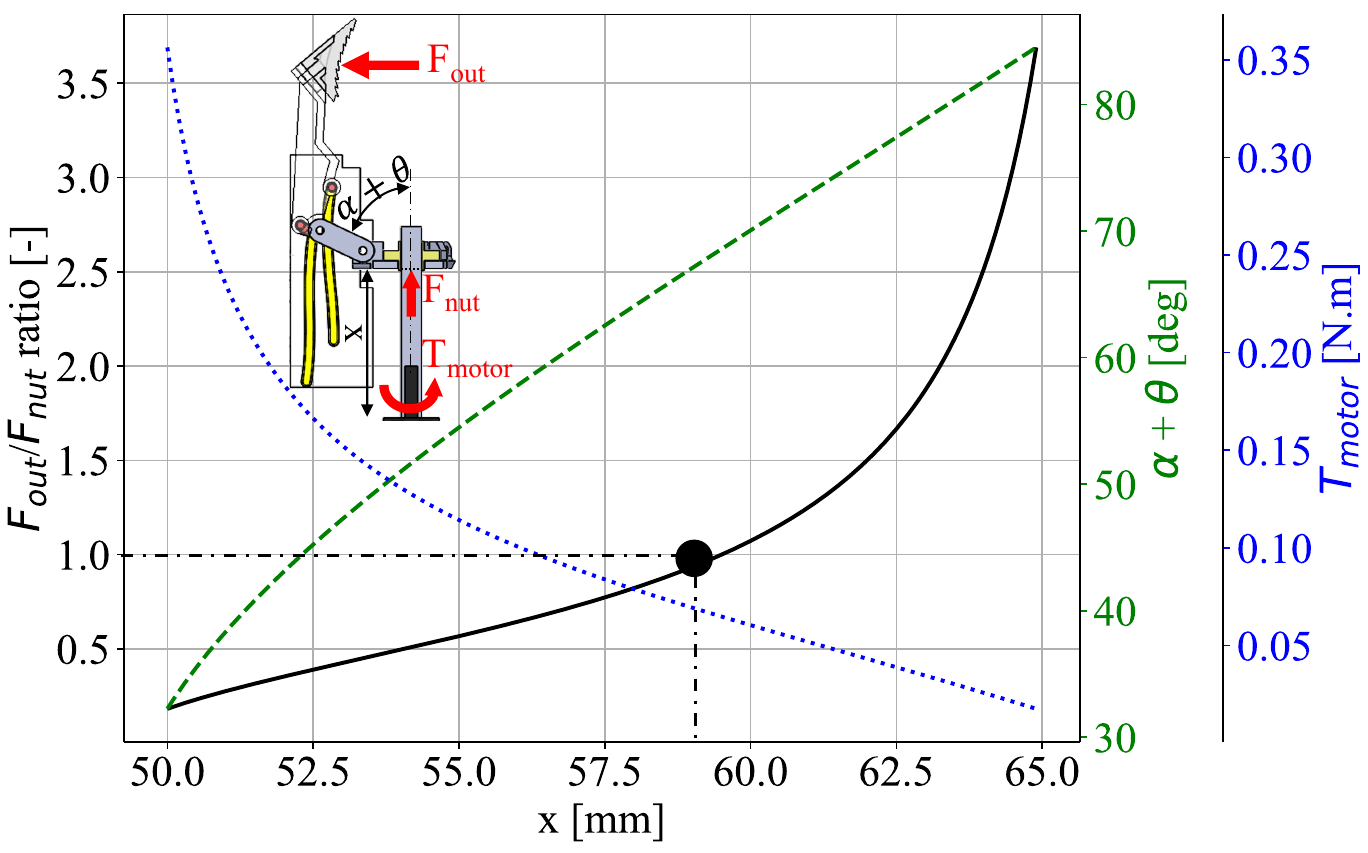}
  \caption{
  Power transmission with respect to linear distance \textit{x} travelled by the lead-screw \textit{nut}. \textit{Continuous} black curve represents force transmission \textit{ratio} between the force normal to the finger pad $F_{out}$ and the thrust force at the lead-screw \textit{nut} $F_{nut}$. We limit \textit{x} to $59$~mm to reach a maximum force ratio of $1:1$. \textit{Dashed} green curve represents the angle $\alpha + \theta$. As this angle approaches $90$\textdegree~ the transmission ratio increases exponentially. \textit{Dotted} blue curve represents the motor torque $T_{motor}$ required to achieve $30$~N $F_{out}$.
  }
  \label{fig:des_plots}
\end{figure}



Fig.\ref{fig:des_plots} shows the force \textit{ratio} (\cref{eq:ratio}) and the angle $(\alpha$+$\theta)$ between \textit{bar} and lead-screw with respect to lead-screw nut displacement \textit{x}. We limit \textit{x} to $59$~mm to reach a maximum force ratio of $1:1$ (black dot in \cref{fig:des_plots}), for two reasons: to remain in the linear range, and to avoid high force ratios (when $\alpha$+$\theta$ is close to $90$\textdegree). Furthermore, \cref{fig:des_plots} shows how the motor torque $T_{motor}$ (\cref{eq:Torque}) decreases with respect to \textit{x}. For a fingerpad force $F_{out}$ of $30$~N, the maximum torque is $0.35$~Nm. We meet this magnitude with a stepper motor (Nema 17 - PHB 42S 34, bipolar with two stacks) that delivers $0.4~Nm$ holding torque.




%% file: 4_Methodology.tex


We evaluated the gripper through a combination of lab experiments and real apple picking trials in a commercial orchard. First, we assessed whether finger normal forces remain lower than the fruit bruising threshold (Bruising test). We then analyzed the grasp strength for each actuation mode (i.e. suction only, fingers only, and both) while varying the pose of the apple w.r.t. the gripper (Grasp strength test). To test for occlusions we performed $34$ fruit picks using our physical proxy \cite{Velasquez2022} with added leaves and apple clusters (Occlusion test). Our real-world trial consisted of picking $25$ apples in a commercial orchard (success rate $> 90\%$).

The fruit picking tests were designed to examine the gripper's performance only and did not include perception, path planning, or other subcomponents that would be required in an integrated system. As described in~\cite{Velasquez2022}, we first define a pick direction $v$ and estimated apple center $p$ (relative to the true apple center and stem direction). The gripper's palm is then placed $5$~cm away from the targeted apple ($p - 5v$), oriented along the pick direction $v$. The gripper is moved along $v$ until either two suction cups engage or the gripper moved 5cm. The fingers were then actuated. The pick motion is a simple pull-back along $-v$, until the fruit is either picked or the grasp fails.

\begin{figure*}[!htp]
  \centering      
  \includegraphics[width=0.95\linewidth]{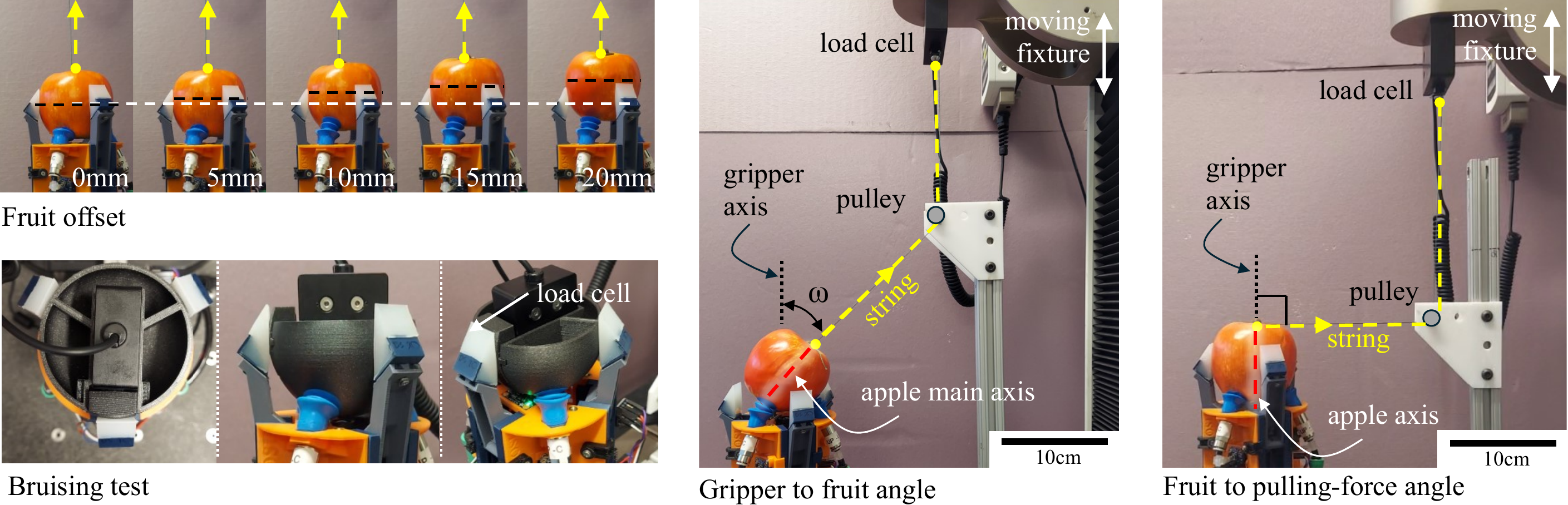}
  \caption{Load cell tests. \textit{Top-left}: Grasp strength test with different fruit offsets. \textit{Bottom-left}: Setup used to measure the normal force exerted by an individual finger. \textit{Middle}: Grasp strength test with the apple's main axis and string aligned, but  varying the gripper alignment. \textit{Right}: Grasp strength test with the apple's main axis aligned with the gripper and the string oriented at $90\deg$, inducing a rotational moment.}
  \label{fig:met_loadCells}
\end{figure*}

\subsection{Bruising test}
The bruising test measures the inward force (i.e. \(F_{out}\)) that the fingers apply to the fruit in the fruit clamping region (\cref{fig:des_gripper}-\textit{Middle}). We measured the inward force at different lead-screw nut displacement values (every $2$~mm from $50$~mm to $60$~mm) to validate the physical model described in \cref{subsub:camfingers}. Fig.~\ref{fig:met_loadCells}~(Bottom left) shows the experimental setup where two fingers were constrained and the third applied force against a load cell (MR03-100). 


\subsection{Grasp strength test}
We measure grasp strength as the force required to pull the apple from the gripper's grasp. This force will vary with the position of the apple in the gripper and the direction of pull from the stem. Our tests capture this variation as follows (see \cref{tab:met_loadcell}):

\begin{itemize}
    \item \textbf{Fruit offset} controls the displacement of the apple from the palm (bigger displacements mean the fingers contact {\em below} the fruit's equator). We varied the starting position of the apple w.r.t. the palm in increments of $5~mm$ from $0~mm$ to $20~mm$ (\cref{fig:met_loadCells}-\textit{Top-left}).
    \item \textbf{Gripper to fruit angle} controls the angle of the tensile forces relative to the orientation of the gripper. The stem is aligned with the pull direction and the gripper is rotated in increments of $15$\textdegree~from $0$\textdegree~to~$45$\textdegree (\cref{fig:met_loadCells}-\textit{Middle}).
    \item \textbf{Rotational pull} evaluates the resistance to {\em rotational} force applied to the fruit. The pull direction is orthogonal to the stem, simulating a force that would rotate the fruit to pull it out of the gripper (\cref{fig:met_loadCells}-\textit{Right}).
\end{itemize}

To conduct the tests we used a materials testing machine (Mark10-ESM1500) with a load cell (MR03-100) attached to the moving fixture and kept the gripper fixed (\cref{fig:met_loadCells}-\textit{Right}). We tied one end of an inextensible fishing line to the load cell and the other end to a $75~mm$ diameter styrofoam apple placed on the gripper's palm. For each trial, we first actuated the gripper to grasp the fruit, and then moved the machine's fixture upward at $100~mm/min$ until the apple was pulled from the grasp. 

We performed the tests with three different actuation modes (suction cups, fingers, or both) in order to capture the effects of each.

\begin{table}[]
\centering     
\begin{threeparttable}
\caption{Force tests performed with a load cell.}
\label{tab:met_loadcell}
\begin{tabular}{|l|r|c|c|c|}
\hline
\multicolumn{1}{|c|}{{Control variable}}                             & \multicolumn{1}{c|}{{Values}} & {Actuation} & {Reps} & {Total} \\ \hline
\begin{tabular}[c]{@{}l@{}}Fruit offset {[}mm{]}\end{tabular}      & 0, 5, 10, 15, 20                     & f                  & 5             & 25             \\ \hline
\begin{tabular}[c]{@{}l@{}}Gripper to fruit\\angle {[}\degree{]}\end{tabular} & 0, 15, 30, 45                        & s, f, s-f            & 5             & 60             \\ \hline
\begin{tabular}[c]{@{}l@{}}Rotational pull {[}\degree{]}\end{tabular}   & 90                                   & s, f, s-f            & 5             & 15             \\ \hline
\end{tabular}
\begin{tablenotes}
 \small
 \item Actuation modes (s: suction, f: fingers, s-f: both)
\end{tablenotes}
\end{threeparttable}
\end{table}

\begin{table}[]
\centering     
\begin{threeparttable}
\centering
\caption{Fruit-pick trials with gripper and UR5e manipulator.}
\label{tab:pick_trials}
\begin{tabular}{|c|l|c|c|}
\hline
\multicolumn{1}{|l|}{Exp.} & Domain & \multicolumn{1}{l|}{Actuation} & \multicolumn{1}{l|}{Trials} \\ \hline
1 & Proxy apple (clusters) & s-f & 9 \\ \hline
2 & Proxy apple (clusters and leaves) & s-f & 25 \\ \hline
3 & \multirow{2}{*}{Commercial orchard} & s & 10 \\ \cline{1-1} \cline{3-4} 
4 &  & s-f & 30 \\ \hline
\end{tabular}
\begin{tablenotes}
 \small
 \item Actuation modes (s: suction, f: fingers, s-f: both)
\end{tablenotes}
\end{threeparttable}
\end{table}




\begin{figure*}[!htp]
      \centering      
      \includegraphics[width=0.95\linewidth]{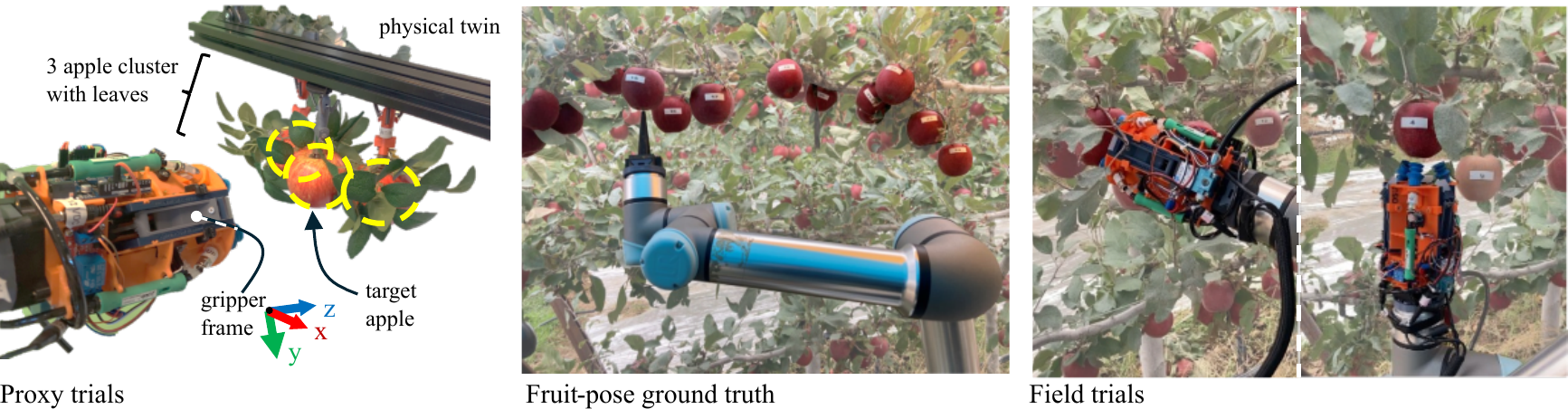}
      \caption{\textit{Left}: Trials using the physical apple proxy from~\cite{Velasquez2022} tuned to the median stem-detachment force and branch stiffness observed during field trials. \textit{Middle}: Procedure used to capture the ground truth of the apple's pose with a probe attached to the wrist of the manipulator. \textit{Right}: Two samples of the apple pick trials in the field. Notice the fruit occlusion.}
      \label{fig:met_picks}
   \end{figure*}


\subsection{Occlusion test}
This test used our physical apple proxy (which mimics the kinematic and dynamic behavior of an apple and stem) from~\cite{Velasquez2022}. We configured the proxy with median values observed during real apple picks ($16~N$ fruit detachment force and $455~N/m$ branch stiffness, as measured from the force/torque sensor on the robot's wrist) (see \cref{fig:met_picks}-\textit{Left}). The fake apples had a $75-mm$ diameter and were filled with sand to match the weight of real ones ($220$\textpm $15~gr$). We evaluated different poses of the gripper w.r.t the apple by varying the gripper's pitch angle every $15$\textdegree~from $0$\textdegree~(underneath the apple's calyx) to $120$\textdegree~(above the fruit's equator), similar to the approach described in~\cite{Velasquez2024}. In this setup (with no occlusion) the gripper can successfully ``pick'' the apple regardless of the pitch angle.

To create the first occlusion test we positioned three apples adjacent to each other so they were touching (picking the middle one). For the second occlusion test we additionally surrounded the middle apple with (fake) leaves. 

\subsection{Field trials}
We evaluated our gripper in a commercial apple orchard (Prosser, WA; variety: `Envy') during the Fall of 2023. These tests were designed to examine the gripper's performance only and did not include perception, path planning, or other subcomponents that would be required in an integrated system. Each targeted apple and its stem were initially probed to measure its ground truth pose as shown in \cref{fig:met_picks}-\textit{Middle}. The pick direction $v$ was set to the direction of the stem (approaching the apple from the bottom). 
Table~\ref{tab:pick_trials} summarizes the apple pick field trial results.


   

%% file: 5_Results.tex



\begin{figure*}[!tph]
  \centering      
  \includegraphics[width=1.0\linewidth]{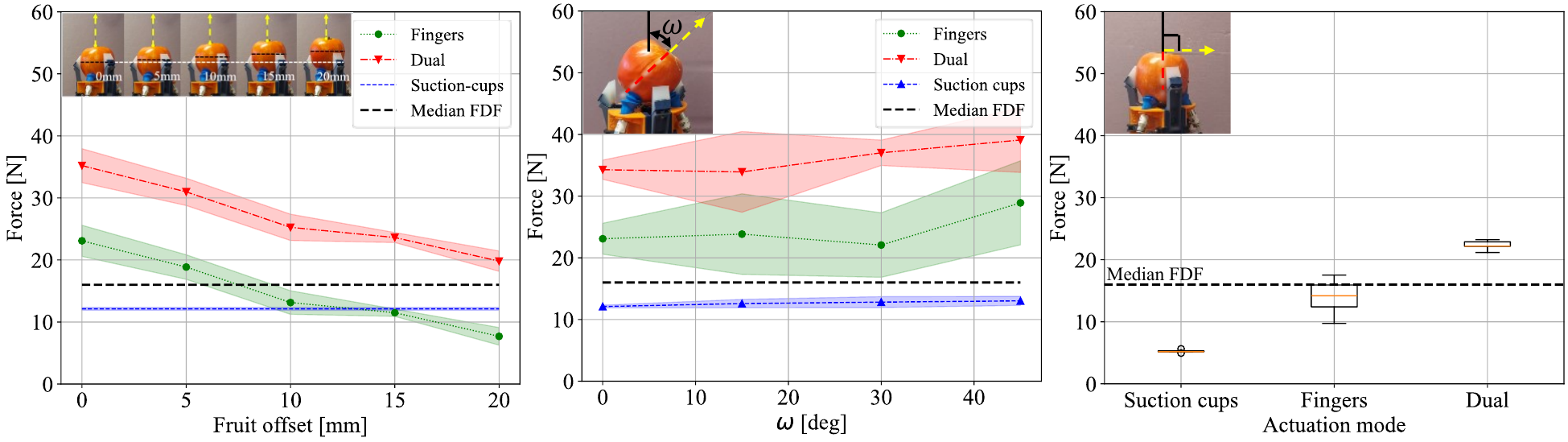}
  \caption{Results from grasp strength tests. \textit{Left}: Grasp strength for fruit offset vs actuation mode. \textit{Middle}: Grasp strength vs angle of pull force. \textit{Right}: Rotational pull results. The median fruit detachment force (FDF) is the dashed black line.}
  \label{fig:res_loadCell}
\end{figure*}

Our experimental results showed that dual actuation, i.e. suction + fingers, significantly increased grasp strength for all test conditions. Additionally, the occlusion trials showed that the telescoping fingers are effective at `pushing' away neighboring fruit in clusters. Finally, during the real apple picking trials, the pick success rate was over $90\%$.

\subsection{Bruising test}
\begin{figure}[!tph]
  \centering      
  \includegraphics[width=0.95\linewidth]{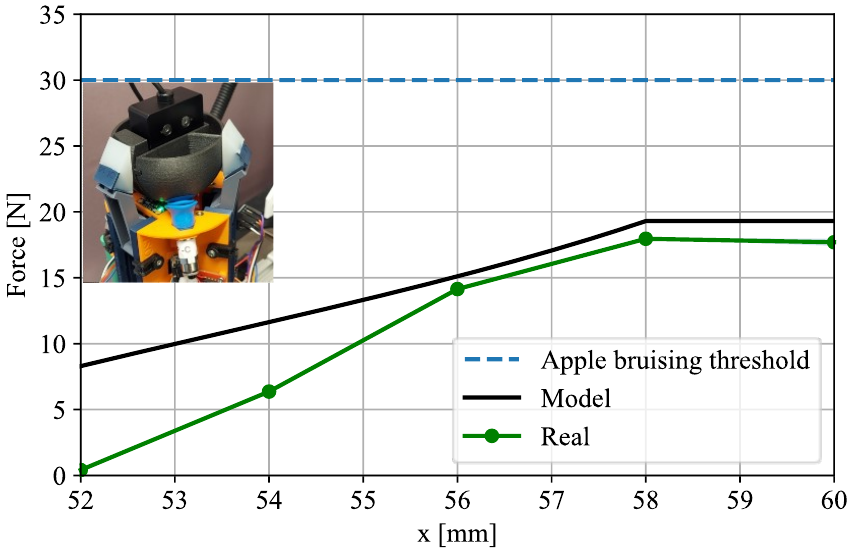}
  \caption{Bruising test. \textit{Green} line shows the load cell measurements.  \textit{Black} line is the predicted $F_{out}$ from the physical model in \cref{subsub:camfingers}. The measured values are similar in shape, but slightly below, the predicted values; both are below the bruising threshold (\textit{Blue dashed} line). The initial difference ($52$~mm to $55$~mm) is likely due to manufacturing tolerances and friction.}
  \label{fig:res_model}
\end{figure}

Fig. \ref{fig:res_model} shows the plot of the force normal to the finger-pad $F_{out}$ versus the lead-screw nut displacement (every $2$~mm from $52$~mm to $60$~mm). The real force $F_{out}$ linearly increases from $0$~N to $18$~N until the lead-screw nut reaches the displacement limit at $58$~mm, similar to the predicted $F_{out}$ from the physical model (derived in \cref{subsub:camfingers}). The difference between the real and predicted force values is likely due to manufacturing tolerances and friction.

\subsection{Grasp strength test}
\textit{Fruit offset}. Fig. \ref{fig:res_loadCell}-\textit{Left} shows the force exerted by the gripper at each offset and with each actuation mode. In suction mode, the grasp strength provided by the three suction cups remains constant at $12~N$, which is lower than the fruit detachment force. At zero offset with the fingers, the grasp strength is above the median fruit detachment force. As expected, the grasp strength with fingers only decreases as the offset increases, from $23.1$\textpm$2.5~N$ (@$0$~mm offset) to $7.7$\textpm$1.4~N$ (@$20$~mm offset). However, with both modes actuated the strength remains above the fruit detachment force even with a $20~mm$ offset.

\textit{Gripper to fruit angle}. Fig. \ref{fig:res_loadCell}-\textit{Middle} shows that the angle of the force w.r.t. to the gripper does not cause a decrease in grasp strength. On the contrary, the grasp strength creeps up from $34.3$\textpm$1.6~N$ (@$0$\textdegree)~to $39.1$\textpm$5.3~N$ (@$45$\textdegree)~in dual mode, and from $23.1$\textpm$2.5~N$ (@$0$\textdegree)~to $28.9$\textpm$6.8~N$ (@$45$\textdegree)~in fingers only mode. We hypothesize that as the angle $\omega$ increases, the normal reaction forces exerted by the fingers will also increase. Regardless, the fingers alone are sufficient to counteract the pulling force.

\textit{Rotational pull}. Fig (\ref{fig:res_loadCell}-\textit{Right}) shows a boxplot for each actuation mode. Again, suction mode alone is insufficient to counteract a rotational pull ($5.25$\textpm$0.2~N$). The suction grasp is also less robust to shear loading, failing at approximately 50\% of the suction strength under tensile loads. Using just fingers brings the grasp strength close to the fruit detachment force ($14$\textpm$2.6~N$). Using both brings the grasp strength above the fruit detachment force to $22.3$\textpm$0.7~N$. 

Interestingly, in general the result with both modes is higher than the sum of each mode actuated individually. We hypothesize that this is because the fingers keep the fruit pressed into the suction cups.

\subsection{Occlusion trials}
For clustered apples (no leaves) we achieved a $100\%$ success rate ($9$ trials). As shown in Fig.~\cref{fig:res_clusters}-\textit{Top}, this is because the wedge-shaped fingers moved the neighboring fruit away. When we added leaves, we succeeded in $24$ of the $25$ trials. In the failed trial, we observed that the leaves interfered with one of the suction cups.


\subsection{Field trials}
A summary of the statistics of fruit properties, \gls{fdf}, branch stiffness, and gripper x-axis offset w.r.t. fruit is shown in \cref{tab:sum_stats}. We first attempted picking apples with just suction cups, then with both. Note that the pick motion was simply a pull-back; no attempt was made to perform a `smarter' pick motion.

\begin{figure*}[!t]
  \centering      
  \includegraphics[width=0.95\linewidth]{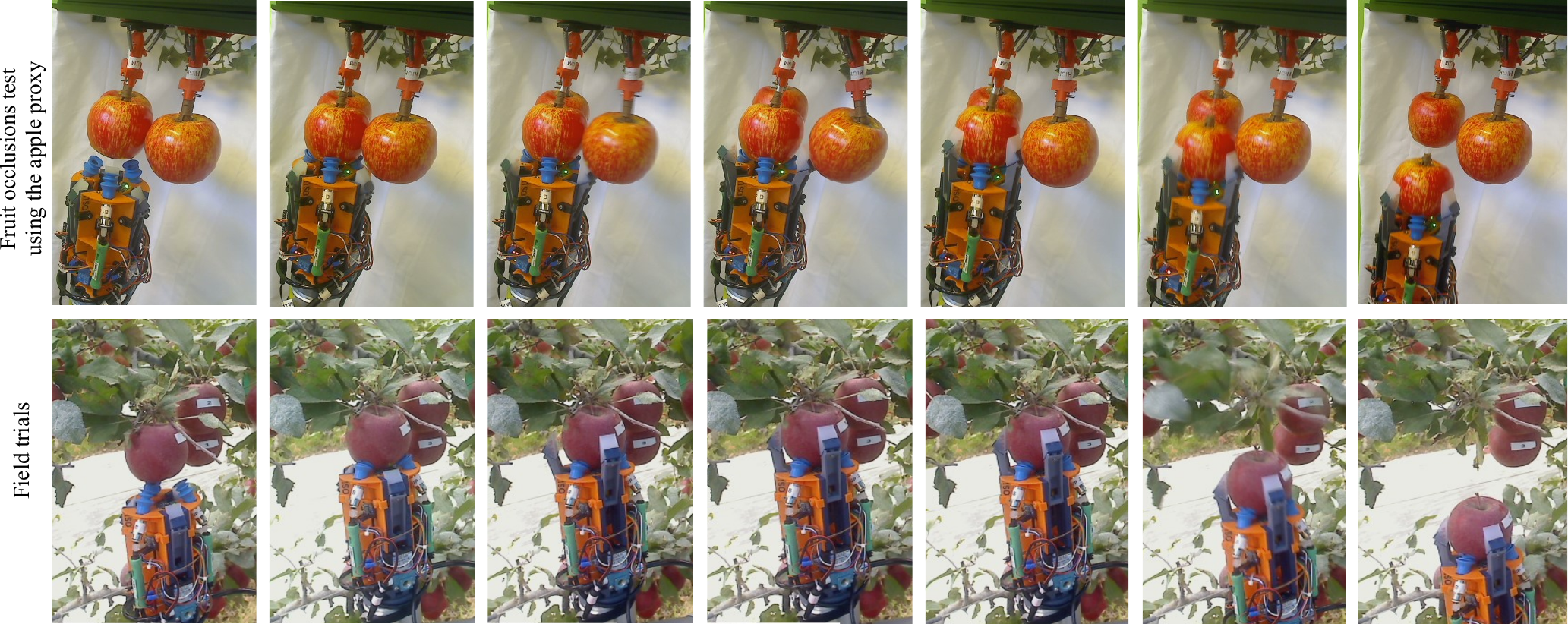}
  \caption{Fruit pick sequence. \textit{Top}: Sequence of the gripper picking the middle apple using the physical proxy. \textit{Bottom}: Example field trial (cluster).}
  \label{fig:res_clusters}
\end{figure*}

\textit{Suction mode}. These trials had good initial grasps, but the suction cups failed before the apple was picked. This was expected, since the grasping force from the suction cups is not sufficient. Just one of our $10$ picks resulted in actually picking the apple.
\textit{Dual mode}. In contrast, in dual mode, we targeted $25$ apples and were able to pick $92\%$. A total of $19$ apples were picked in the $1st$ attempt, two apples in the $2nd$ attempt, and one in the $3rd$ attempt. The median \gls{fdf} was $15$~N close to our initial reference values; however, values up to $38$~N were observed (\cref{tab:sum_stats}). A time sequence of one of the field trials is shown in \cref{fig:res_clusters}-\textit{Bottom}.



%

\begin{table}[]
\caption{Field trials statistics summary}
\label{tab:sum_stats}
\resizebox{\linewidth}{!}{%
\begin{tabular}{|l|c|c|c|c|c|c|}
\hline
Variable                      & \begin{tabular}[c]{@{}c@{}}Number\end{tabular} & Min & Q1  & \textbf{Median} & Q3  & Max  \\ \hline
Fruit diameter {[}mm{]}       & 24                                                    & 70  & 76  & \textbf{78}     & 81  & 86   \\ \hline
Fruit height {[}mm{]}         & 24                                                    & 61  & 70  & \textbf{73}     & 75  & 79   \\ \hline
Fruit weight {[}gr{]}         & 21                                                    & 181 & 222 & \textbf{235}    & 248 & 284  \\ \hline
Net FDF {[}N{]}               & 22                                                    & 7   & 11  & \textbf{15}     & 28  & 38   \\ \hline
Tangential FDF {[}N{]}        & 22                                                    & 1   & 3   & \textbf{7}      & 19  & 31   \\ \hline
Normal FDF {[}N{]}            & 22                                                    & -2  & 7   & \textbf{12}     & 19  & 33   \\ \hline
Branch stiffness {[}N/m{]}    & 34                                                    & 71  & 234 & \textbf{410}    & 780 & 1324 \\ \hline
Gripper-fruit offset {[}mm{]} & 39                                                    & 1   & 5   & \textbf{10}     & 16  & 30   \\ \hline
\end{tabular}
}
\end{table}


%% file: 6_Conclusion.tex



In this study, we developed and evaluated a compact tandem-actuated gripper designed for selective fruit harvesting, featuring both suction and cam-driven finger actuation. The gripper first employs suction to gently seize the fruit, followed by the use of cam-driven fingers to displace obstacles and secure the fruit, ensuring it can withstand the forces encountered during the pick. Testing with a materials testing machine demonstrated a maximum grasping strength of up to $40$~N under various fruit-gripper poses. Additionally, our tests with a physical apple proxy confirmed the gripper's robustness in handling fruit clusters and leaves, achieving a grasp success rate over $96\%$. Field validation in a commercial apple orchard further confirmed the gripper's efficacy, with $24$ out of $25$ targeted apples successfully picked, predominantly on the first attempt.
Future work will focus on implementing an air-pressure servoing controller to enhance the engagement of the suction cups and optimizing the cam-tracks to accommodate a wider range of fruit diameters.